\pdfoutput=1
\documentclass[11pt]{article}
\usepackage{amssymb}
\usepackage[final]{acl}

\usepackage{times}
\usepackage{latexsym}

\usepackage[T1]{fontenc}
\usepackage[utf8]{inputenc}

\usepackage{microtype}

\usepackage{inconsolata}

\usepackage{graphicx}
\usepackage{tikz}
\def\checkmark{\tikz\fill[scale=0.4](0,.35) -- (.25,0) -- (1,.7) -- (.25,.15) -- cycle;}

\usepackage{graphicx}
\graphicspath{ {./images/} }
\usepackage{amsmath,amssymb}
\usepackage{multirow}
\usepackage{caption}
\usepackage{subcaption}
\usepackage{algorithm} 
\usepackage[noend]{algpseudocode}
\usepackage{arydshln}
\usepackage{booktabs}
\usepackage{multirow}
\usepackage{fdsymbol}
\usepackage{siunitx}
\usepackage{dirtytalk}
\usepackage{amsmath}

\usepackage{array}
\usepackage{xspace}

\newcolumntype{L}[1]{>{\raggedright\let\newline\\\arraybackslash\hspace{0pt}}m{#1}}
\newcolumntype{C}[1]{>{\centering\let\newline\\\arraybackslash\hspace{0pt}}m{#1}}
\newcolumntype{R}[1]{>{\raggedleft\let\newline\\\arraybackslash\hspace{0pt}}m{#1}}

\makeatletter
\def\adl@drawiv#1#2#3{%
        \hskip.5\tabcolsep
        \xleaders#3{#2.5\@tempdimb #1{1}#2.5\@tempdimb}%
                #2\z@ plus1fil minus1fil\relax
        \hskip.5\tabcolsep}
\newcommand{\cdashlinelr}[1]{%
  \noalign{\vskip\aboverulesep
           \global\let\@dashdrawstore\adl@draw
           \global\let\adl@draw\adl@drawiv}
  \cdashline{#1}
  \noalign{\global\let\adl@draw\@dashdrawstore
           \vskip\belowrulesep}}
\makeatother

\makeatletter
\DeclareRobustCommand\onedot{\futurelet\@let@token\@onedot}
\def\@onedot{\ifx\@let@token.\else.\null\fi\xspace}

\makeatother

\newcommand{\modelours}{\textsc{AdpLVR}\xspace}

\newcommand{\modelbertdeb}{\textsc{BERT-NLI}\xspace}
\newcommand{\modeladapternli}{\textsc{AdpNLI}\xspace}

\newcommand{\modelbert}{\textsc{BERT-Base}\xspace}

\newcommand{\modelroberta}{\textsc{RoBERTa-Base}\xspace}
\newcommand{\modelfine}{\textsc{Ft}\xspace}
\newcommand{\modelfinedeb}{\textsc{FtAdv}\xspace}
\newcommand{\modeladapter}{\textsc{Adp}\xspace}
\newcommand{\modeladapterdeb}{\textsc{AdpAdv}\xspace}
\newcommand{\modeladaptermmd}{\textsc{AdpMMD}\xspace}

\title{Unlabeled Debiasing in Downstream Tasks via Class-wise \mbox{Low Variance Regularization}}

\author{Shahed Masoudian$^1$, Markus Frohmann$^{1,2}$, Navid Rekabsaz$^{3}$, Markus Schedl$^{1,2}$ \\
  $^1$ Johannes Kepler University Linz, Austria \\
  $^2$ Linz Institute of Technology, AI Lab, Austria \\
  $^3$ Thomson Reuters Labs, Zug, Switzerland \\
  \texttt{shahed.masoudian@jku.at} 
  }

\begin{document}
\maketitle
\begin{abstract}
% Language models more than often reflect societal biases present in their training data.
% While some unsupervised methods have been developed to debias sentence embeddings of language models, fine-tuning them with new datasets can reintroduce biases back to the model.
% Recently studies have proposed various debiasing techniques, such as mutual information removal, contrastive learning, and adversarial training to reduce the protected attribute information from the embeddings of the language model.
% A common limitation of these methods is their reliance on labeled protected attributes to remove biases from model embeddings.
% In this work we demonstrate that debiased language models can regain biases upon fine-tuning with new data and introduce a new regularization method designed to reduce attribute information in embeddings without requiring access to attribute labels.

% Pre-trained 
Language models frequently inherit societal biases from their training data. Numerous techniques have been proposed to mitigate these biases during both the pre-training and fine-tuning stages.
However, fine-tuning a pre-trained debiased language model on a downstream task can reintroduce biases into the model.
Additionally, existing debiasing methods for downstream tasks either (i) require labels of protected attributes (e.g., age, race, or political views) that are often not available or (ii) rely on indicators of bias, which restricts their applicability to gender debiasing since they rely on gender-specific words.
To address this, we introduce a novel debiasing regularization technique based on the class-wise variance of embeddings.
Crucially, our method does not require attribute labels and targets \textit{any} attribute, thus addressing the shortcomings of existing debiasing methods. 
Our experiments on encoder language models and three datasets demonstrate that our method outperforms existing strong debiasing baselines that rely on target attribute labels while maintaining performance on the target task.\footnote{The code for the experiments is available on GitHub: \url{https://github.com/ShawMask/UnlabeledDebiasing}}

% The datasets used in this fine-tuning stage typically do not feature labels corresponding to a given sensitive attribute.
% This makes using current bias mitigation methods impractical in such cases since they all share one commonality: reliance on labels.

% \emph{without requiring access to attribute labels}.
% It regularizes [X some concise info about the algorithm X].
% Compared to existing debiasing methods relying on attribute labels, our experiments using two commonly used encoder LMs and three datasets show that [our method's name] improves debiasing performance compare to baselines of the same caliber while preserving task performance.

% last sentence needs more precision.

\end{abstract}

\section{Introduction and Background}
\label{sec:introduction}

Language Models (LMs) based on encoders are used for a variety of purposes such as document classification~\cite{founta:hatespeech, de:bios}, job recommendation~\cite{kumar:fairness}, text generation~\cite{ronen:tinystories}, or as text encoder for multimodal models such as text-to-audio~\cite{liu:audioldm} or text-to-image~\cite{bahani:t2i} models.
% Encoded inside the model, exist societal biases that are rooted in the dataset.
These models often encode societal biases rooted in the corpora used for training~\cite{mehrabi-bias,rekabsaz:biasbert}, which causes a distributional shift of embeddings, hence affecting their outputs either with disproportionate misclassification of documents belonging to minority groups or unfair ranking of the documents~\cite{rekabsaz_2021_societal,melchiorre_gender_fairness}.

Several works focus on reducing the effect of these biases by improving model performance related to some specific fairness metric (\textit{empirical fairness}) or by making the model blind to the existence of a certain attribute (\textit{representational fairness})~\citep{shen-representational}.
For instance, \citet{shen-representational} leverage contrastive learning to improve empirical fairness. Recent works focus mostly on efficiency and user flexibility when it comes to debiasing using modular approaches such as sub-networks or adapters~\cite{houlsby_2019_param_efficient}.
\citet{hauzenberger2023parameter} introduce a modular debiasing scheme with adversarial training~\cite{elazar_2018_adv} and mutual information reduction~\cite{colombo_2021_novel} to control the bias in encoder LMs.
\citet{kumar2023parameter} use adversarial training with adapters~\cite{pfeiffer2021adapterfusion} to improve representational fairness on document classification.
Finally, \citet{masoudian:congater} used gated adapters to improve representational fairness while preserving task performance for classification and retrieval tasks. % for classification and information retrieval tasks.

Although these methods effectively reduce sensitive attribute information and enhance fairness~\cite{zerveas_2022_mitigating,shen-representational} through blindness, they depend on attribute labels to align the distribution of the target attribute.
Since the user input data contains numerous nuanced protected attributes, such as age, race, religion, etc., it is challenging to collect labeled data for each individual attribute across every task.
Moreover, supervised debiasing methods typically require training on each attribute individually, scaling linearly with the number of attributes. This complexity highlights the need for more efficient and scalable approaches to handle multiple protected attributes in debiasing efforts.

To address this limitation, some works attempt to debias language models without using attribute labels.
\citet{zhou:debias-sentence} employ contrastive learning combined with instance weighting to reduce the bias encoded in the language model.
Moreover, \citet{cheng_2020_fairfil} utilize post-hoc contrastive learning to enhance the fairness of pre-trained encoder language models concerning gender bias.
\citet{ghanbarzade:gendertuning} integrate the masking objective used during the pre-training of encoder language models with fine-tuning on gender-specific tasks to address gender bias.

% To address this shortcoming, some studies attempt to debias LMs without using attribute labels. 
% \citet{zhou:debias-sentence} uses contrastive learning in combination with instance weighting to reduce the bias encoded in the language model. 
% Moreover, \citet{cheng_2020_fairfil} apply post-hoc contrastive learning to improve the fairness of pre-trained encoder LMs with regard to gender bias.
% \citet{ghanbarzade:gendertuning} integrates the masking objective used during pre-training of encoder LMs with fine-tuning on gender-specific tasks to tackle gender bias.
% These methods, while effective at addressing gender bias when explicit gender indicators are present, have the advantage of not relying on labeled data.

These methods address gender bias without requiring labeled data using explicit gender indicators present within the text.
However, they are ineffective against other biases such as age, race, or political view of the user, as well as implicit gender bias when gender information is removed from the text, limiting their possible use cases.

In this work, we bridge this gap by introducing a new \textit{regularization scheme based on class-wise variance} to reduce unknown (unlabeled) representational bias in the embeddings of LMs. Our regularization enforces low-variance embeddings, which results in mitigating any possible distributional shift caused by unknown attributes in the model's embeddings. With this method we force the model to produce robust embeddings that are informative about the classification task but contain less information about the protected attributes resulting in fair representation of the protected attributes. This gives our method the advantage of not relying on any type of information on the attribute during debiasing.
To the best of our knowledge, we are the first to address the debiasing of \textit{arbitrary} attributes without having access to attribute labels.
%, which is the main contribution of our work.

We demonstrate the effectiveness of our method on document classification taskss using adapters~\cite{pfeiffer2021adapterfusion} and two commonly used encoder LMs, \modelbert and \modelroberta.
Furthermore, we show that our method, when compared to existing \textit{supervised} debiasing methods, can enhance attribute removal while still showing competitive classification task performance. %on three datasets and four attributes.

% \paragraph{Contributions.} \textbf{1.)} We are introducing a novel regularization scheme that allows the model to produce robust embeddings that contain less information about arbitrary attribute unrelated to the main task \textit{without} any access to target attributes. 
% \textbf{2.)} We take advantage of the batch centers during optimization of the model as additional loss signal during the training to maintain task performance while debiasing the language model. We also show that our method outperforms previous baselines with regard to attribute information removal while competing in task performance.

% \vspace{-2mm}
\section{Problem Formulation}
\label{subsec:pf}
% \vspace{-2mm}

In recent years, adapter networks~\cite{houlsby_2019_param_efficient,pfeiffer2021adapterfusion} have emerged as an efficient way of training models on downstream tasks. In addition to their improved training efficiency, adapters keep the backbone LM weights fixed, helping preserve information within the model.

In our initial study, we assess how much gender information can be extracted from commonly used encoder LMs. We use adapters~\citep[][Pfeiffer variation]{pfeiffer2021adapterfusion} in combination with \citep[][\modelbert]{devlin-etal-2019-bert} and, additionally, a gender-debiased version of the same model~\citep[][\modelbertdeb]{wu:bert-nli} debiased for empirical fairness, on two downstream classification tasks.
We then train probes on the embeddings of the fine-tuned models to check how much information about gender can be extracted from both model variations and report the average balanced accuracy as indicator of gender information in the embeddings. 

Table~\ref{tab:debiased} shows the result of both models for occupation prediction (BIOS~\citep{de:bios}) and mention prediction (PAN16~\citep{rangel:pan16}) datasets.
We observed that task performance of the debiased LM, \modelbertdeb, is consistently lower than \modelbert, which aligns with observations by \citet{ghanbarzade:gendertuning}.
Moreover, training adapters contain the gender information to a great extent on BIOS; while on PAN16, \modelbertdeb leaks more gender information in the embeddings, although it has already been subject to debiasing.
% in an earlier stage. 
% Based on this observation, we infer that debiased models are still prone to bias even when their weights are frozen.
% hence we focus our attention on downstream task where model is being used and no further training is done .
\begin{table}[t]
\caption{Result of adapter training of BERT (\modelbert) and a gender-debiased version of BERT (\modelbertdeb) on two datasets with gender as the protected attribute. Here, \textit{Task} corresponds to accuracy on the main classification task, and \textit{Gender} is the balanced accuracy of the model concerning the protected attribute.
}
\vspace{-1mm}
\label{tab:debiased}
\scalebox{0.83}{
\begin{tabular}{l|ll|ll}
\toprule
\multirow{2}{*}{Model} & \multicolumn{2}{c}{\textbf{BIOS}} & \multicolumn{2}{c}{\textbf{PAN16}} \\ 
 & \textbf{Task}$\uparrow$ & \textbf{Gender}$\downarrow$ & \textbf{Task}$\uparrow$ & \textbf{Gender}$\downarrow$ \\ \midrule
\textsc{\modelbert} & $84.3_{0.1}$ & $67.0_{0.1}$ & $92.4_{0.1}$ & $70.7_{0.1}$ \\ \midrule
\textsc{\modelbertdeb} & $84.1_{0.1}$  & $64.5_{0.1}$ &$88.2_{0.1}$  & $73.7_{0.1}$  \\ 
\bottomrule
\end{tabular}
}
\vspace{-2mm}
\end{table}

This provides strong motivation for using debiasing methods during fine-tuning, even when using an already debiased pre-trained LM.
However, as surveyed in §~\ref{sec:introduction}, existing debiasing methods either rely on attribute labels or are limited to attributes with explicit indicators in the text, such as gender. Furthermore, there exists a plethora of sensitive attributes, and labeling them all is challenging across tasks. This increase in number also affects debiasing complexity as it scales with the number of attributes.
Thus, a method that addresses this gap would be highly desirable.
In the following, we outline how we solve this gap.

We start with our problem setting, formulated as follows:
Given a set of $N$ documents with $k$ classes, we are interested in having robust high-dimensional embeddings ($Z \in \mathbb{R}^d$) for document classification which are (i) informative about the classes but (ii) contain as little information as possible about any arbitrary protected attribute ($\rho$) not directly related to the classification task.
% 
% Most existing debiasing methods either take advantage of attribute labels to align the embedding space~\cite{elazar_2018_adv,masoudian:congater,kumar2023parameter} or use indicators of bias inside the text and leverage them to only gender-debias the LM~\cite{cheng_2020_fairfil, zmigrod-etal-2019-counterfactual} 
Our approach to debiasing deviates from existing ones in two crucial ways:
(i) It is independent of labeled attributes, and (ii) it targets \emph{any} protected attribute simultaneously.

\section{Low Variance Regularization (LVR)}
\label{sec:methodology}

We formulate our regularization scheme based on $k$ centers, each representing a class in the dataset with $d$ dimension $\{C_1,C_2,...C_k | C_i \in \mathbb{R}^d\}$, where $d$ is the model's embedding size. We aim to adjust the parameters of the network if the variance of the embeddings in a batch is high, which intuitively results in the mitigation of any undesirable distributional shift that might exist in the embeddings. 
% Considering that we have $k$ classes, class-wise variance can be a good estimation of this regularization loss. 
Since we have $k$ classes, class-wise variance is a good proxy for this regularization loss. 

We define the regularization loss as the distance between embeddings ($Z \in  \mathbb{R}^d$) of class $i$ in a given batch from their corresponding center.
For each batch, we calculate the center of embeddings that belong to the same class ($C_i$), which results in $k$ centers. To account for noisy data points and empty batches, we use the weighted sum of the current batch center $C_i^b$ and the normalized weighted sum of previous batch centers $C_i^{b-1}$ where $\omega$ is a hyperparameter to control the influence of previous batch and found through grid search. 
% Equation~\ref{eq:center} shows how the centers for each batch of data are calculated
The centers are calculated as follows:

\begin{equation}
% \vspace{-1mm}
% \label{eq:center}
    C_i^b = (1 - \omega)\frac{Z_1 + Z_2 ... Z_m}{m} +\omega C_i^{b-1},
% \vspace{-1mm}
\end{equation}

% % ~\ref{eq:center} formulates how the centers of each class is calculated as the average of all dimensions.
% We then calculate the center of each class as the average of all dimensions:

% \begin{equation}
% \label{eq:center}
%     C_i = \frac{X_1 + X_2 ... X_m}{m}
% \end{equation}

\noindent
where $m$ is the number of samples for the $i^{th}$ class in a batch. In practice, if there are no samples of a class within a batch, we ignore it; and if only one sample of the class is in the batch, the center becomes the sample itself.
% % and if there are two or more samples, we calculate and estimate the center using~\ref{eq:center}.
We then define the regularization loss as the sum of distances for each specific sample belonging to class $i$ from the estimated center of the batch: 

% To account for noisy samples in some batches, we also memorize the centers of previous batches and then calculate a weighted combination $\omega$ of the current and the previous batch to smoothen the regularization loss.
% In addition, we use the centers as new classification samples, classifying them into one of $k$ classes.
% Then, the loss calculated by the centers is reintroduced to the model as $\mathcal{L}_c$.
% \vspace{1mm}
% \[
% \resizebox{.95\hsize}{!}{
% \begin{cases}
%     \mathcal{L}_r = \sum_{i=1}^{m}\sqrt{\sum_{j=1}^{d}(x_j^i-c_j)^2},& \text{if } x_j^i\geq c_j\\
%     \mathcal{L}_r = 0,              & \text{otherwise}\\
% \end{cases}
% }
% \]

\begin{equation}
    \mathcal{L}_r = \sum_{i=1}^{k}\sum_{r=1}^{m}\sqrt{\sum_{j=1}^{d}(z_{jr}^i-c_j^i)^2},
\end{equation}
% \vspace{1mm}

\noindent
where $c_j^i$ is the center value for the $j^{th}$ dimension of the $i^{th}$ class and $z_{jr}^{i}$ is the value for the $j^{th}$ dimension of the $r^{th}$ embedding for the $i^{th}$ class. This corresponds to reducing the class-wise variance of the embeddings created by the model, which in turn reduces distributional shift that might exist in the data points of the same class and results in the alignment of the embeddings. We also use the calculated centers as extra input for the classification task and calculate the loss of the centers. 
We show later in §~\ref{sec:results} that this added loss term is essential to mitigate degradation in task performance.
% Thus the overall loss is a linear combination and is achieved from equation~\ref{eq:loss}. 

\noindent 
The overall loss then becomes a linear combination:

\begin{equation}
% \vspace{-1mm}
\label{eq:loss}
    \mathcal{L}_{total} = \mathcal{L}_t + \lambda \mathcal{L}_r +\mathcal{L}_c
% \vspace{-1mm}
\end{equation}

where $\mathcal{L}_t$ is the classification loss, $\mathcal{L}_r$ is the regularization loss, and $\mathcal{L}_c$ is the loss to classify the calculated centers belonging to each class.

% \vspace{-2mm}
\section{Experimental Setup}
\label{sec:setup}
% \vspace{-2mm}

% As mentiond in section~\ref{sec:introduction}, 
For our experiments, we follow previous works and focus on transformer-based language models. We use \modelbert and \modelroberta, in combination with adapters~\citep{pfeiffer2021adapterfusion} for each task. 
Trained in this way, we denote models using our debiasing method as \textbf{\modelours}.

We use the following document classification datasets:
occupation prediction (\textbf{BIOS};~\citet{de:bios}) with gender as protected attribute,
hate speech detection (\textbf{FCDL18};~\citet{founta:hatespeech}) with race for protected attribute,
and mention detection (\textbf{PAN16};~\citet{rangel:pan16}), corresponding to a multi-attribute setting with age and gender as protected attributes.
For each dataset, we remove all explicit indicators of protected attributes following previous works~\citep{hauzenberger2023parameter,kumar2023parameter,masoudian:congater} from the text.
% to reduce the effect of the input distribution on the model bias to only 
% ; instead, we report balanced accuracy of the probes to address this imbalance in the dataset. 

\begin{table*}[ht]
\centering
\vspace{-6mm}
\caption{Results of debiasing using \modelbert and \modelroberta.
\modelours has no access to labeled attribute while \modelfinedeb, \modeladapterdeb and \modeladaptermmd  have access to the attribute information in the form of attribute labels.}
\vspace{-2mm}
\small
\scalebox{1}{
\begin{tabular}{l l ll | ll | ll | ll}
\toprule
\multirow{2}{*}{Model} &\multirow{2}{*}{Type} &\multicolumn{2}{c}{\textbf{BIOS}} & \multicolumn{2}{c}{\textbf{FDCL18}}& \multicolumn{2}{c}{\textbf{PAN16-Gender}}& \multicolumn{2}{c}{\textbf{PAN16-Age}} \\
& & \textbf{Task}$\uparrow$ & \textbf{Probe}$\downarrow$ & \textbf{Task}$\uparrow$ & \textbf{Probe}$\downarrow$&  \textbf{Task}$\uparrow$ & \textbf{Probe}$\downarrow$&  \textbf{Task}$\uparrow$ & \textbf{Probe}$\downarrow$ \\\midrule

\multirow{6}{*}{\modelbert}
&\modelfine &$84.6_{0.4}$ &$67.3_{0.8}$ & $81.0_{1.0}$ & $92.9_{1.8}$ &$93.6_{1.8}$ &$69.6_{0.8}$ &$93.6_{1.8}$ &$42.3_{0.9}$ \\
&\modeladapter &$84.3_{0.1}$ &$67.0_{0.1}$ &$80.0_{0.1}$  &$93.3_{0.4}$ &$92.4_{0.1}$ &$70.7_{0.1}$ &$92.4_{0.1}$ &$42.4_{0.}$ \\
&\modeladapternli &$84.1_{0.1}$ &$64.5_{0.1}$ &$81.2_{0.6}$  &$93.5_{0.6}$ &$88.2_{0.1}$ &$73.7_{0.1}$ &$88.2_{0.1}$ &$42.5_{0.1}$ \\
&\modelfinedeb &$84.0_{0.3}$ &$60.8_{0.2}$ & $81.0_{1.0}$ &$84.4_{4.0}$ &$92.4_{0.8}$ &$59.8_{0.7}$ &$92.4_{0.8}$ &$31.3_{1.1}$ \\
&\modeladapterdeb  &$84.2_{0.1}$ & $61.9_{0.5}$&$79.8_{0.3}$  &$75.6_{0.5}$ &$92.2_{0.1}$ &$\textbf{54.2}_{0.4}$ &$92.1_{0.1}$ &$\textbf{21.7}_{0.1}$ \\
&\modeladaptermmd  & $84.4_{0.2}$ &$65.3_{0.3}$ & $80.1_{0.2}$ &$81._{0.3}$ & $91.4_{0.4}$ &$67.4_{0.3}$ & $92.0_{0.8}$ &$36.8_{0.7}$ \\ \cdashlinelr{2-10}
&\textbf{\modelours}  &$84.0_{0.2}$ &$\textbf{59.2}_{0.3}$ &$81.7_{0.1}$ &$\textbf{66.7}_{0.9}$ & $91.3_{0.1}$ &$54.4_{0.1}$ & $91.3_{0.1}$ &$21.9_{0.2}$ \\
\midrule

\multirow{6}{*}{\modelroberta}
&\modelfine &$84.5_{0.4}$ &$66.2_{0.7}$ &$80.6_{0.4}$ &$93.2_{1.2}$ &$98.5_{0.1}$ &$63.6_{0.4}$ & $98.5_{0.1}$ &$22.7_{0.8}$ \\
&\modeladapter &$84.3_{0.1}$ &$67.3_{0.7}$ &$80.0_{0.6}$ &$94.0_{0.6}$ &$98.2_{0.1}$ &$62.8_{0.4}$ &$98.1_{0.1}$ & $31.9_{0.1}$ \\
&\modelfinedeb &$84.1_{0.3}$ &$61.6_{0.3}$ &$80.5_{1.0}$ &$83.6_{1.9}$ &$98.2_{0.1}$ &$52.0_{0.9}$ &$98.2_{0.1}$ & $24.1_{1.4}$ \\
&\modeladapterdeb  &$84.0_{0.1}$ &$62.9_{0.1}$ &$80.0_{0.5}$ &$79.7_{0.3}$ &$98.1_{0.1}$ &$53.7_{0.7}$ & $98.0_{0.1}$ &$22.3_{1.0}$  \\
&\modeladaptermmd  & $84.3_{0.2}$ &$64.2_{0.3}$ & $80.0_{0.1}$ &$80._{0.5}$ & $97.8_{0.1}$ &$60.4_{0.3}$ & $98.0_{0.4}$ &$27.1_{0.3}$ \\ \cdashlinelr{2-10}
&\textbf{\modelours}  &$83.8_{0.1}$ &$\textbf{55.6}_{0.3}$ &$81.5_{0.2}$ &$\textbf{77.3}_{0.1}$ & $97.7_{0.1}$ &$\textbf{51.1}_{0.4}$ & $97.7_{0.1}$ &$\textbf{20.6}_{0.8}$ \\
% \midrule
    
% \multirow{6}{*}{\modelmini}
% &\modelfine &\textbf{82.4}$_{0.1}$ &$65.7_{0.2}$ &$79.9_{1.0}$ &$92.4_{0.6}$ &$91.5_{0.1}$ &$65.4_{0.8}$ &\textbf{91.5}$_{0.1}$ &$40.9_{0.4}$ \\
% &\modeladapter &$82.1_{0.1}$ &$65.2_{0.4}$ &$81.3_{0.1}$ &$93.5_{0.8}$ &$81.5_{0.1}$  &$65.5_{0.4}$  &$81.6_{0.1}$  & $37.4_{1.4}$  \\
% &\modelfinedeb &$81.7_{0.1}$ & $60.4_{0.4}$ &$79.3_{0.8}$ &$81.4_{2.1}$ &$90.3_{0.4}$ &$58.6_{0.8}$ &$90.3_{0.4}$ &$27.9_{1.8}$ \\
% &\modeladapterdeb  &$82.1_{0.2}$ &$61.4_{0.6}$ &$80.7_{0.1}$ &$75.9_{1.2}$ &$81.3_{0.1}$  &$53.1_{0.1}$  &$81.1_{0.1}$  & $21.7_{0.3}$  \\
% &\modeladaptermmd  & & & & & & & & \\
% &\modelours  & $81.5_{0.2}$ &$77.3_{0.1}$ &$79.9_{0.2}$ &$81.7_{0.1}$ & & & & \\
\bottomrule

\end{tabular}
}

\label{tab:results:all_models}
\vspace{4mm}
\end{table*}

% \vspace{-4mm}

\begin{table*}[ht]
\centering
\caption{Results for adapter based training of \modelbert with different memory ($\omega$) and center loss ($\mathcal{L}_c$) combinations.}
\small
% \scalebox{1}{
\begin{tabular}{l l ll | ll | ll | ll}
\toprule
\multirow{2}{*}{$\omega$} &\multirow{2}{*}{$\mathcal{L}_c$} &\multicolumn{2}{c}{\textbf{BIOS}} & \multicolumn{2}{c}{\textbf{FDCL18}}& \multicolumn{2}{c}{\textbf{PAN16-Gender}}& \multicolumn{2}{c}{\textbf{PAN16-Age}} \\
& & \textbf{Task}$\uparrow$ & \textbf{Probe}$\downarrow$ & \textbf{Task}$\uparrow$ & \textbf{Probe}$\downarrow$&  \textbf{Task}$\uparrow$ & \textbf{Probe}$\downarrow$&  \textbf{Task}$\uparrow$ & \textbf{Probe}$\downarrow$ \\\midrule

% \multirow{6}{*}{\modelbert}
- & - & $80.2_{0.8}$ &$60.9_{0.3}$ & $81.1_{0.1}$ &$72.8_{0.2}$ & $91.2_{0.1}$ &$54.4_{0.3}$ & $91.2_{0.1}$ &$24.3_{0.1}$ \\
$\checkmark$ & - & $83.5_{0.2}$ & $59.6_{0.1}$  &$81.1_{0.2}$ &$72.0_{0.3}$ &  $91.1_{0.1}$ &$54.4_{0.1}$ & $91.1_{0.1}$ &$22.3_{0.1}$ \\
-&$\checkmark$ &$83.8_{0.4}$ & $59.8_{0.5}$ & $81.5_{0.2}$ &$70.9_{0.8}$ & $91.4_{0.2}$ &$55.5_{0.3}$ & $91.4_{0.2}$ &$23.1_{0.1}$ \\
\multicolumn{2}{c}{\modelours} &  $84.0_{0.3}$ & $59.2_{0.2}$ & $81.7_{0.1}$ &$66.7_{0.9}$ & $91.3_{0.1}$ &$54.4_{0.1}$ & $91.3_{0.1}$ &$21.9_{0.2}$ \\
\bottomrule
\end{tabular}
% }

\label{tab:results:ablation}
\vspace{3mm}
\end{table*}

\paragraph{Baselines.}
We choose baselines as follows:
\textbf{\modelfine}, \textbf{\modeladapter} and \textbf{\modeladapternli} as fine-tuned versions of the entire model and adapter-based training of the model and adapter-based training for BERT model trained on debiased NLI, respectively, without using any additional bias mitigation method.
We also select recent in-process debiasing algorithms as strong baselines, relying either on adversarial training~\citep[][\textbf{\modelfinedeb}; \textbf{\modeladapterdeb}]{elazar_2018_adv} or mutual information reduction~\citep[][\textbf{\modeladaptermmd}]{colombo_2021_novel} to reduce the bias encoded within the embeddings and increase representational fairness.
% We select additional baselines as follows: 
Note that all supervised methods use labels of the target attribute to align the embeddings, while \modelours does \textit{not} have access to any attribute label throughout the training.

\paragraph{Training.}
We follow the setup of previous works using the same datasets~\citep{hauzenberger2023parameter,kumar2023parameter,masoudian:congater}.
Specifically, we use a maximum of 120 tokens for the BIOS dataset and 40 tokens for FDCL18 and PAN16 since they comprise comparatively short tweets.
We train each model for 15 epochs with a learning rate of $2 \times 10^{-5}$.
We select reduction factors for adapters on BIOS, PAN16, and FCDL18 as $2$, $1$, and $2$, respectively, as they led to the best task performance. %during hyperparamer tuning
% Since the effect of each loss term on model optimization is different,
Since each loss term affects each model differently,
we train baselines with a fixed $\lambda=1$ for supervised debiasing and our unsupervised \modelours with $\lambda=0.1$. We also select $\omega=0.3$ as it performed best across all datasets in our grid search.
%as it showed in our grid search it works best for all datasets. % as we observed that our regularization loss is strong and can easily overtake task training.

We train five probes consisting of two-layer fully connected networks and \texttt{tanh} activation function for 40 epochs and a learning rate of $1 \times 10^{-4}$ to predict protected attributes from embeddings~\cite{hauzenberger2023parameter}.

To evaluate task performance, we use accuracy as the evaluation metric. In order to evaluate the performance of bias mitigation methods we use balanced accuracy of the probes. We choose balanced accuracy to account for any unbalanced dataset with regard to the distribution of the protected attributes. For gender and dialect attributes, if balanced accuracy is around $50\%$ it shows that the model is confusing the protected attribute; for age this value should be close to $20\%$ because there are five classes for age.  
Furthermore, we run our experiments three times for each model and report the mean and standard deviation of three runs to account for variations in the training process. 

% Table~\ref{tab:results} show the result of training of all 3 datasets. As we can see from the table, Our regularization method is able to remove information from the embeddings of the model without accessing the labeled attribute while introducing some task performance loss. Having no labeled data makes it difficult for the model to understand the different distributions that exists in the dataset which results in enforcing the embedding a bit more than when you already have a defined distribution for different attributes.

% \vspace{-2mm}
\section{Results and Discussion}
\label{sec:results}
% \vspace{-2mm}
% Before delving into the results we would like to reiterate that BIOS and FCDL18 contain only one target attribute respectively gender and dialect and Pan16 dataset contain 2 attributes namely gender and age. 
% Baseline models had access to target attribute during training and are specifically trained to reduce bias of the target attribute while \modelours does not have access to any attribute information during training.

Table~\ref{tab:results:all_models} shows the task and probe performance of the baselines and \modelours. 

In our single attribute experiments on FCDL18 and BIOS, using both \modelbert and  \modelroberta, \modelours is able to remove information about protected attributes considerably better than \textit{all} the baselines on BIOS and FCDL18. As for task performance, we observe a decrease in accuracy with \modelours compared to the baselines on BIOS. 
Remarkably, our regularization method even shows an improvement in task performance on FCDL18, demonstrating the robustness of its embeddings. 

In our multi-attribute experiment on PAN16 (2 protected attributes), we observe that \modelours performs slightly worse than the best-performing model, \modeladapterdeb, on the main task. However, unlike \modeladapterdeb, which has access to the protected attribute label during training, \modelours crucially does not rely on attribute labels for bias mitigation; yet, it still outperforms the baselines in protected attribute information removal. Overall, we observe that, with \modelbert, \modelours shows slightly higher balanced accuracy compared to \modeladapterdeb for both protected attributes, while with \modelroberta, \modelours similarly shows improved mitigation performance.

% This finding is in line with previous work on \emph{supervised} debiasing, 
Notably, other debiasing methods show similar decreases in task performance. Still, on FCDL18, \modelours clearly outperforms all \textit{supervised} baselines on the main task and information removal.
% This finding is in line with recent findings of \citet{masoudian:congater}.

% but still task performance is influence by debiasing on Pan16 and BIOS dataset whereas the same improvement of the task performance can be seen on FCDL18.

% all in all we cannot consider the slight task performance decrease as a downside to our method as any other baseline method is also experiencing it which is also reported by~\cite{masoudian:congater}.

\paragraph{Ablation Study.}
To ensure all parts of our method are necessary to achieve its performance, we conduct an ablation study where we remove (i) the memory of the previous batch, $\omega$, and (ii) the center loss $\mathcal{L}_c$ introduced in section~\ref{sec:methodology}.
Table~\ref{tab:results:ablation} shows the result of this ablation study.
By removing $\omega$, the balanced accuracy of the probe considerably increases, meaning that the robustness of the embeddings toward protected attributes is reduced.
Thus, more information about unknown, unrelated attributes influences the final output of the model to a larger extent.

Moreover, we observe that task performance clearly degrades when removing $\mathcal{L}_c$.
Overall, the best-performing model, both in terms of task performance and probe balanced accuracy, is the one that has both $\omega$ acting as memory of previous batches for the model and $\mathcal{L}_c$, corresponding to our class-center-based loss.
% This holds across all datasets and attributes, showing that all components of our method are crucial.

% We can see the results of the baseline models and \modelours on Bios dataset were the target attribute is gender and for FCDL18 which the target attribute is dialect of the user. In both cases we can see that \modelours is able to remove attribute information better than baseline models with minimal harm to the task. For each case the baseline models used the labeled information of the attribute to align the embedding of transformer model, in contrast \modelours didn't have access to any information from the embeddings. We can see that in both datasets our method is able to reduce the target information from the embeddings considerably better than of those with access to labeled embeddings. \textbf{Multi Attribute}: In Table~\ref{tab:results:all_models} we can see the results for Pan16 dataset where target attribute are gender and age. As we can see from the table the results show that all the baseline which had access to labeled data are reducing the attribute information from embedding of the model less than our proposed method. This further shows that our method is able to remove any attribute information regardless of the how they exist in the dataset without looking at them during training. 

% \vspace{-2mm}
\section{Conclusion}
% \vspace{-2mm}
In this work, we focus on representational fairness and introduce a novel regularization and optimization scheme to debias encoder LMs without accessing protected attribute labels.
We show the effectiveness of our method using two encoder LMs across three datasets and multiple protected attributes. We demonstrate that our method enhances debiasing while maintaining task performance compared to strong baselines. To the best of our knowledge, our method is the first that can mitigate bias of any \emph{arbitrary} target attribute by generating robust embeddings best suited for the classification task. Since our method does not rely on attribute labels, we hope it paves the future for more accessible, effective, and efficient debiasing of encoder-based transformer models.

% We showed that in comparison to strong baselines that use attribute information our method is able to outperform them in reducing attribute information and still maintain a competitive main task performance. 

% We introduced a generic regularization method based on the estimation of class centers during the training of a Transformer encoder.
% It acts as a novel \textit{unsupervised} debiasing method to reduce attribute information from its embeddings.
% In our experiments across three datasets and four attributes, we show that by regularizing networks with the memory of old batches and adding another loss term to guide task learning, our method effectively removes attribute information from the embeddings with only minimal harm to task performance.
% Our method even outperforms supervised baselines having access to attribute information during training and debasing.
% Since our method does not rely on labeled attribute information in any form, we hope it paves the future for more accessible, effective, and efficient debiasing of encoder-based Transformer models.

% \vspace{-3mm}
\section*{Limitations}
One limitation of this work is the definition of gender used in all datasets, which is limited to binary female/male, lacking an inclusive and nuanced definition of gender.
Moreover, although our method proved independent of attribute labels, a thorough evaluation would require more datasets with a variety of defined attributes. 
Another limitation of this work is the task in which we narrowed our study to classification tasks. We acknowledge that the findings of this paper might not be applicable to other tasks such as retrieval or recommendation. Furthermore, our study is focused on transformer-based language models which put an additional limitation on the generalization of the work to other models such as CNNs or LSTM-based language models.
Due to the lack of suitable datasets, we relied on datasets commonly used in the debiasing literature. In FDCL18, race is restricted to \emph{African American} and \emph{White American}, which does not reflect real-life scenarios.
Furthermore, we follow previous works ~\cite{sap_2019_risk,ravfogel_2020_null, zhang_2021_disentangling} and use labels of protected attributes assigned using another model, making them not fully representational of the real data distribution. 
A final limitation of this work is the lack of suitable datasets for multi-attribute settings, in which we could demonstrate that our approach can handle even more attributes than demonstrated with PAN16 simultaneously.
% Pan16 dataset which is used in this paper only contains two attributes and is not a strong indicator of a multi-attribute scenario. 

% Do not forget about this.

% \section*{Acknowledgments}
\section*{Acknowledgements}
% \vspace{-2mm}
This research was funded in whole or in part by the Austrian Science Fund (FWF): \url{https://doi.org/10.55776/P33526}, \url{https://doi.org/10.55776/DFH23}, \url{https://doi.org/10.55776/COE12}, \url{https://doi.org/10.55776/P36413}.

% Bibliography entries for the entire Anthology, followed by custom entries
%\bibliography{anthology,custom}
% Custom bibliography entries only
\bibliography{references}

% \appendix

% \section{Example Appendix}
% \label{sec:appendix}

% This is an appendix.

\end{document}